# Deep Learning for Flight Demand Forecasting

Liya Wang,[1] Amy Mykityshyn,[2] Craig Johnson,[3]
*The MITRE Corporation, McLean, VA, 22102, United States*

Benjamin D. Marple,[4]
*Federal Aviation Administration*

**Inspired by the success of deep learning (DL) in natural language processing (NLP), we applied cutting-edge DL techniques to predict flight departure demand in a strategic time horizon (4 hours or longer). This work was conducted in support of a MITRE-developed mobile application, Pacer, which displays predicted departure demand to general aviation (GA) flight operators so they can have better situation awareness of the potential for departure delays during busy periods. Field demonstrations involving Pacer's previously designed rule-based prediction method showed that the prediction accuracy of departure demand still has room for improvement. This research strives to improve prediction accuracy from two key aspects: better data sources and robust forecasting algorithms. We leveraged two data sources, Aviation System Performance Metrics (ASPM) and System Wide Information Management (SWIM), as our input. We then trained forecasting models with DL techniques of sequence to sequence (seq2seq) and seq2seq with attention. The case study has shown that our seq2seq with attention performs best among four forecasting algorithms tested. In addition, with better data sources, seq2seq with attention can reduce mean squared error (mse) over 60%, compared to the classical autoregressive (AR) forecasting method.**

## I. Nomenclature

| | | |
|---|---|---|
| $T$ | = | time interval between two consecutive time steps, 15 minutes |
| $t$ | = | time step |
| $p$ | = | number of time lags |
| $\tau$ | = | steps from current time t |
| $\tau_{max}$ | = | maximum look ahead times |
| $X_t$ | = | observed inputs at time t |
| $F_t$ | = | future known inputs at time t |
| $y_t$ | = | response variable at time t |
| $mae$ | = | mean absolute error |
| $mse$ | = | mean squared error |

## II. Introduction

Airport surface congestion and flight delays due to sudden surges in General Aviation (GA) flight demand have been a concern for airlines and the Federal Aviation Administration (FAA). Surface demand is comprised of aircraft from scheduled commercial flights and unscheduled GA operations. Commercial flights are scheduled months in advance and their schedule is shared among the airlines, the airports, and the FAA, whereas GA flights schedules are often more flexible, and, therefore, less predictable. For airports that experience large increases in GA traffic due to special public events (e.g., professional sports games) or airports that have a consistently high percentage of GA traffic (e.g., Van Nuys Airport (VNY), Teterboro Airport (TEB)), it can be quite challenging for FAA traffic managers to strategically manage expected demand.

---

[1] Artificial Intelligence, Lead, Department of Operation Performance
[2] Aviation Systems Engineering, Lead, Department of Safety Intelligence Concepts and Evolution
[3] Group Leader, Department of NAS Future Vision & Research
[4] Operations Research Analyst, Surveillance Branch, ANG-C52

The MITRE Corporation has developed Pacer [1], a mobile application, to support the FAA for efficient surface management. Pacer displays future departure demand information to GA pilots and allows pilots to provide updates to their intended departure times; this information is subsequently used to improve future departure demand predictions. Field demonstrations involving Pacer's previously designed rule-based prediction method showed that the prediction accuracy of departure demand still has room for improvement. Therefore, this research is one of several efforts that is exploring ways to improve prediction accuracy for GA departure demand.

The latest trend in forecasting techniques has shifted from rule-based to intelligent deep learning (DL) methods. Machine learning (ML) methods are different from traditional rule-based methods as shown in Fig. 1. In traditional rule-based methods, the domain experts define the rules, and then the computer makes predictions according to those rules. However, in ML, the rules are learned from data and applied to make predictions. DL is a subset of machine learning that utilizes programmable neural networks.

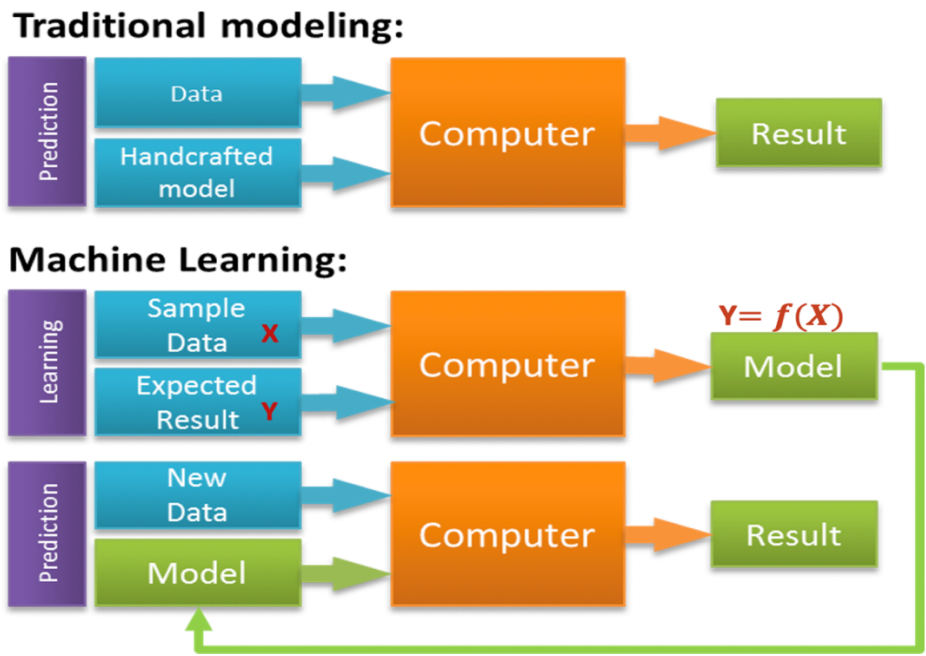

**Fig. 1 Traditional programming vs. machine learning methods comparison**

Rule-based methods have some obvious deficiencies. For example, when the situation is extremely complicated (e.g., image classification for various types of cats vs. dogs), humans often cannot explicitly enumerate all the rules. Table 1 lists the shortcomings of Pacer's rule-based methods and the benefits DL can bring to address those gaps.

**Table 1 Comparison of rule-based and DL methods**

| Deficiencies of Rule-based Methods | Promises of Proposed DL Methods |
|---|---|
| • Rules may not accurately describe reality<br>• Rules may not be complete<br>• No learning ability<br>• Rules cannot be easily extended to other airports<br>• Few factors have been considered | • Rules will be automatically learned from historical data<br>• Rules can be adaptively upgraded with latest available training data<br>• Models can be adaptively extended to other airports with transfer learning<br>• Multiple factors such as hour, day of week, and month can be included |

Aviation flight demand prediction is a real-world application of a multi-horizon time series forecasting problem, which presents challenges due to multiple heterogeneous input variables, including observed past ground information and future known information such as hour of day, day of week, and month, etc. To tackle these complicated problems, our research explores DL forecasting techniques of sequence to sequence (seq2seq) [2] and seq2seq with attention [3] [4], which have demonstrated success in the NLP field for solving the machine translation problems. Those algorithms are designed to extrapolate past sequences to predict future sequences.

With the promise of DL algorithms, aviation researchers are eager to adopt these cutting-edge techniques to address aviation-specific challenges. For example, researchers have applied these techniques to predict unstable approach risk at a specific distance location to the runway threshold [5], air travel demand for a representative city pair [6], flight delays [7], and taxi-out times at Charlotte Airport [8].

The remainder of this paper is organized as follows: Section III gives a short introduction about our input data sources used in the research; and Section IV describes our problem formulations. Section V provides details of the deep learning modeling process, and the results are shown in Section VI. Finally, we conclude and describe next steps in Section VII.

## III. Data Source

DL algorithms run on data and the need for large amounts data is greater than ever. But more than large amounts of data, good data quality is also crucial to get the desired end result. Therefore, we took great effort to prepare our modeling dataset.

For our departure demand prediction, we leveraged two aviation datasets: 1) Aviation System Performance Metrics (ASPM) [9] and 2) System-Wide Information Management (SWIM) [11]. A short description of these two data sources is provided.

### A. Aviation System Performance Metrics (ASPM) dataset

The ASPM dataset is a well-known operational data source that the FAA maintains. It integrates multiple data sources together and provides information about flights arriving or departing an airport. The ASPM dataset has quarter-hour aggregated demand, delays, runway configuration, and weather information needed for forecasting. Table 2 lists the key data items used in our study [10]. In particular, quarter-hour departure demand (DepDemand) is our targeted forecasting objective, and the other elements are our inputs. One shortcoming with ASPM is its long updating frequency, only once per day. In addition, once updated, only the previous day's data are available. Therefore, ASPM is not well suited for frequent forecasting updates.

**Table 2 Data Items from ASPM**

| Column Name | Description |
|---|---|
| Slice_Start_UTC | 15-minute period start time in UTC time |
| Hour | UTC hour (0-23), derived from Slice_Start_UTC |
| Qtr | UTC quarter hour (1 to 4), derived from Slice_Start_UTC |
| Day of week | UTC day of week (1 to 7), derived from Slice_Start_UTC |
| Month | UTC month (1 to 12), derived from Slice_Start_UTC |
| DepDemand | Number of aircraft intending to depart for the period |

### B. System-Wide Information Management (SWIM) dataset

To complement ASPM's deficiency, we brought in another FAA data source, SWIM, which can give us first-hand, near real-time information about the state of an airport's surface. SWIM is updated every minute and has information such as arrival/departure lists and taxi-in/taxi-out times. Therefore, our modeler can reduce prediction error by 1) using shorter look ahead times, and 2) updating the forecasts more frequently.

Compared to ASPM, SWIM data is still in an early development stage. The data is provided in XML format, and then each user group processes it according to their research needs. The version of SWIM data available to support our work is noisy. Therefore, we spent a great effort to clean the data with departure track assembly, data exploration, data visualization, and domain experts' evaluation.

In addition, the SWIM data covers only the core 30 airports [12]. Therefore, we plan to build two types of models for airports: one that is capable of using both ASPM and SWIM data as inputs, and the other that just uses ASPM data when SWIM data are not available.

## IV. Problem Formulation

Flight departure demand forecasting problems can be formulated as multivariate multi-step time series forecasting problems (Eq. (1)). Time lag ($p$) was learned from the data and the maximum look ahead time, $\tau_{max}$, is determined by the data updating frequency and Pacer's needs. Fig. 2 explicitly demonstrates the relationship of inputs and outputs in a given time horizon. The inputs are made up of two parts: past observed inputs and future known inputs. For our targeted departure demand (DepDemand) prediction, Table 3 lists the detailed inputs and outputs. Next, we will discuss how deep learning methods, seq2seq and seq2seq with attention are built to find function $f$.

| Equations | No. |
|---|---|
| $y_{t+\tau} = f(y_{t-1}, y_{t-1}, y_{t-2}, \ldots, y_{t-p}, X_{t-1}, X_{t-2}, \ldots, X_{t-p}, F_{t+1}, F_{t+2}, \ldots, F_{t+\tau})$ | (1) |

where $\tau = [1, \ldots, \tau_{max}]$

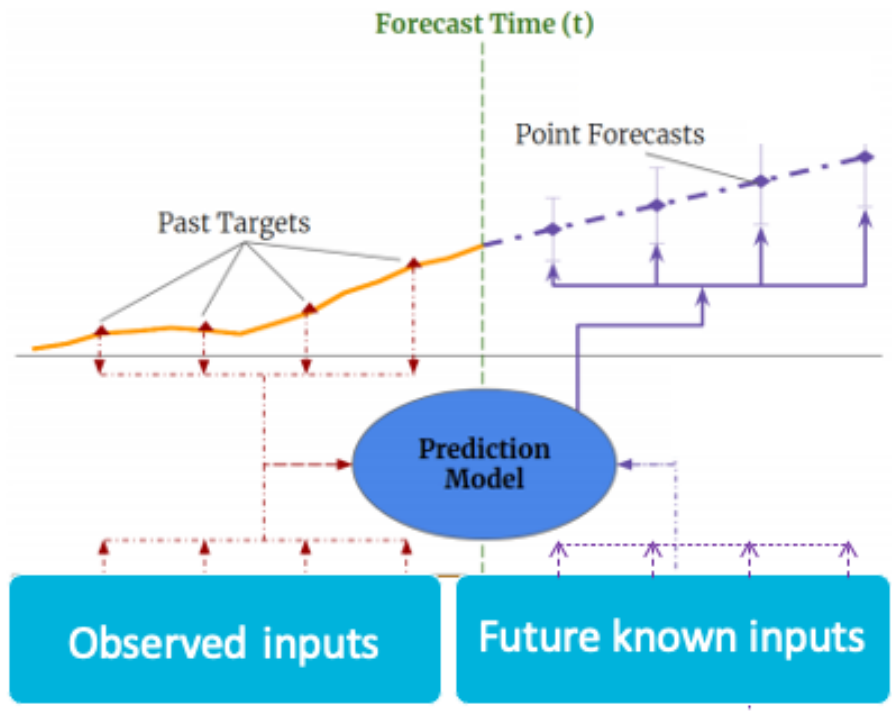

**Fig. 2 Illustration of multi-horizon forecasting with observed inputs and future known inputs ( [13])**

**Table 3 Listed inputs and outputs for departure demand prediction**

| Target output | • DepDemand (from ASPM) |
|---|---|
| **Observed inputs** | • Observed number of departure flights (from SWIM) |
| | • y's history (from ASPM) |
| **Future known inputs** | • Hour of day (from ASPM) |
| | • Quarter of hour (from ASPM) |
| | • Day of week (from ASPM) |
| | • Month (from ASPM) |

## V. Deep Learning for Time Series Forecasting

Solving real-world time series forecasting problems can be challenging. The challenges are multifaceted and include, for example, heterogeneous input variables, accounting for future known factors (e.g., day of week), and the need to perform the same type of prediction for multiple physical sites, like different airports, in our case [14].

This research solves the complicated time series forecasting problems with DL methods, which can bring many benefits. For example, they can handle multiple exogenous variables with complex dependencies, identify the multifaceted relationship between input variables and output variables, learn and adapt, and extend to multiple sites with ease. Our research sought to enable DL to predict aviation flight departure demand for Pacer. Fig. 3 depicts the designed DL model training architecture for Pacer's demand forecasting.

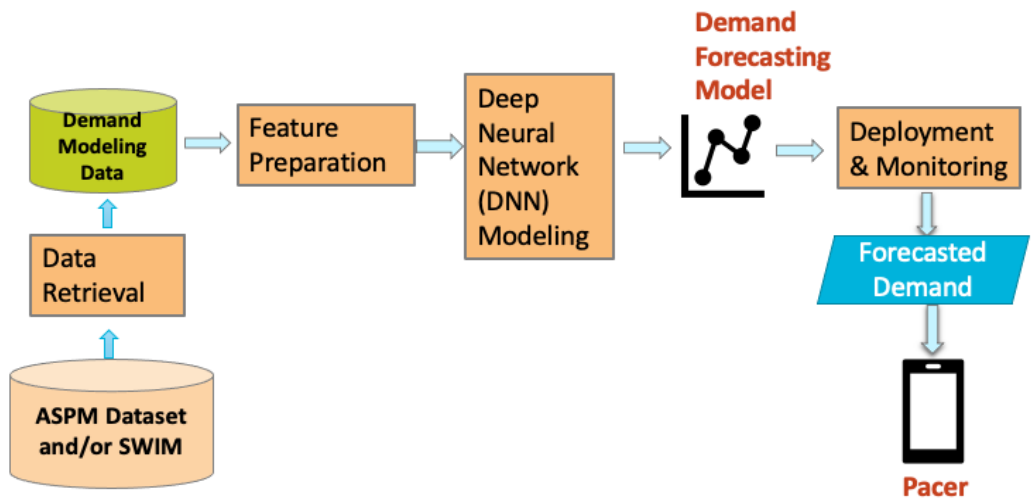

**Fig. 3 Deep learning architecture for Pacer's demand and delay forecasting**

### C. Feature Preparation

When we train DL models, feature preparation plays an essential role. Simply stated, feature preparation transforms raw data into the DL algorithm-required format, as well as improves the model performance. Domain knowledge and feature engineering knowledge are important components at this step. We designed a five-stage feature processing procedure (Fig. 4) to transform raw time series data into sequence data required by our sequence to sequence (seq2seq) and seq2seq with attention models, which are described in the following section.

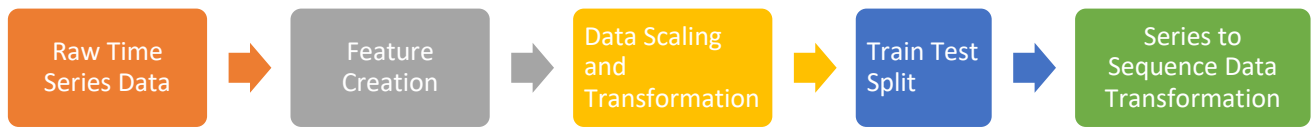

**Fig. 4 Feature processing procedure for Pacer demand forecasting**

### D. Sequence to sequence (seq2seq) Method

Seq2seq was originally designed by Google for machine translation problems and has achieved a lot of success in tasks like machine translation. One famous example is Google Translate, which started using such a model in production in late 2016. These models can be found in two pioneering papers ( [2], [15]).

Inspired by the success of seq2seq in the NLP field, researchers have explored seq2seq for the time series forecasting problems because they have a similar problem structure (e.g., [16]). Seq2seq can achieve better performance than traditional statistical time series forecasting methods, which usually require strong restrictions and assumptions.

As the name suggests, seq2seq takes an input sequence (e.g., sentences in English) and generates an output sequence (e.g., the same sentences translated to Chinese). It does so by using a recurrent neural network (RNN). Two commonly used types of RNN are Long Short-Term Memory (LSTM) and Gated Recurrent Units (GRU) because LSTM and GRU can better handle the vanishing gradient problem [17], which can cause the neural network to not be trained properly.

Seq2seq is made up of two parts: the *encoder* and the *decoder*. It is sometimes referred to as the **Encoder-Decoder Network** (Fig. 5).

- **Encoder:** Uses deep neural network layers and converts the input sequence to a corresponding hidden vector as an initial state to the first recurrent layer of the decoder part.

- **Decoder:** Takes the input from a) the hidden vector generated by the encoder, b) its own hidden states, c) its own current output, and d) future known factors $F$ to produce the next hidden vector and finally predict the next output.

In our research, out future known inputs are categorical variables such as hour of day, day of week, month etc. Therefore, we use an embedding layer to encode them first and then send them to LSTM cells. Because LSTM type of cells performs better than GRU, we chose to proceed with LSTM in our seq2seq model building.

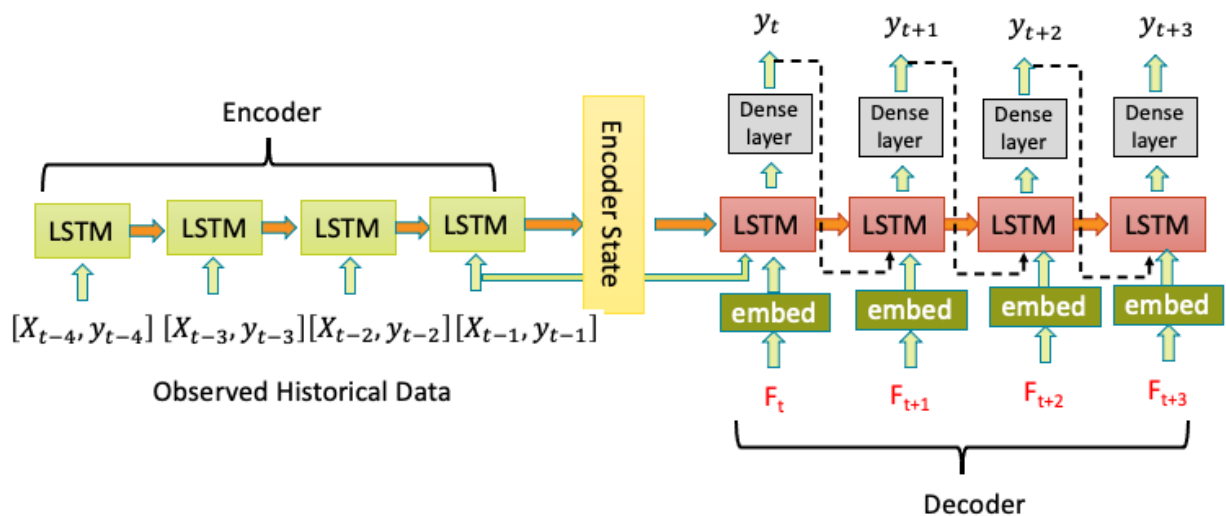

**Fig. 5 Seq2seq architecture for our departure demand time series forecasting**

One of the drawbacks of seq2seq models is that they have difficulty handling long input sequences because their encoding context is wrapped up in a specified size vector. In the case of long sequences, there is a high probability that the initial context has been lost by the end of the sequence.

E. **Sequence to sequence (seq2seq) with Attention Method**

To deal with the aforementioned issue, seq2seq with attention models have been introduced in two papers( [3], [4]). The method introduced a technique called "Attention," which flexibly focuses on different parts of the input sequence at every stage of the output sequence by preserving all the context. Fig. 6 illustrates our customized seq2seq with Luong's attention network structure for departure demand forecasting.

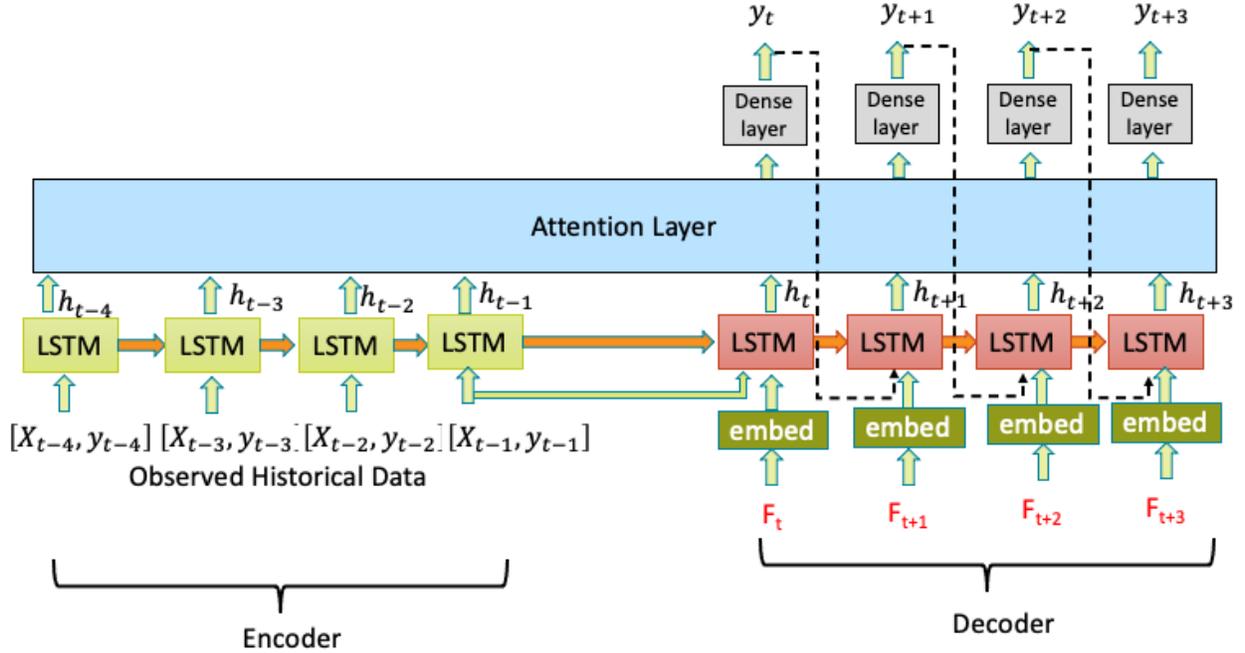

**Fig. 6 Seq2seq with attention architecture for our departure demand time series forecasting**

## VI. Results

This section presents our modeling results with different forecasting methods. We selected Las Vegas McCarran International Airport (LAS) for case study because LAS represents a busy airport with many airline operations as well as a significant number of GA operations that can surge around certain events. To better understand how DL would improve forecasting performance, we also chose two baseline algorithms, linear regression (LR) [18] and the autoregression (AR) [19], for the comparison. The following sub-sections give a short introduction to these baseline algorithms and all models' prediction results.

### A. Baseline Algorithms
### A.1 Linear Regression

In statistics, linear regression is used to model the relationship between a scalar response and one or more explanatory variables. The very simplest case of a single scalar predictor variable *x* and a single scalar response variable *y* is known as simple linear regression (see Fig. 7). The extension to multiple and/or vector-valued predictor variables (denoted with *X*) is known as multiple linear regression, also known as multivariable linear regression. Nearly all real-world regression models are multiple regression models. Note, however, that in these cases the response variable *y* is still a scalar. Another term, multivariate linear regression, refers to cases where *y* is a vector [18].

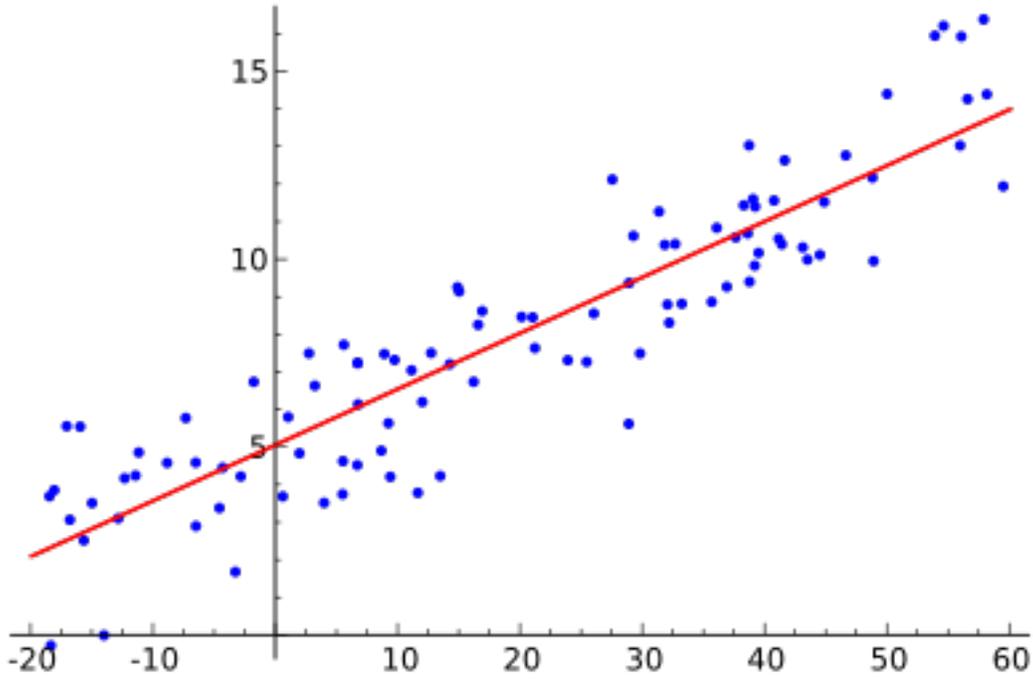

Fig. 7 Example of simple linear regression, which has one independent variable ( [18])

### A.2 Autoregression (AR)

Autoregression (AR) is another popular stochastic process model designed for time series forecasting. The autoregressive model specifies that the output variable depends linearly on its own previous values and on a stochastic term, as shown in Eq. (2). AR methods cannot consider as many factor variables as DL and LR do. The only prior knowledge required is its history.

| Equation | No. |
|---|---|
| $y_t = c + \phi_1 y_{t-1} + \phi_2 y_{t-2} + \cdots + \phi_p y_{t-p} + \epsilon_t$ | (2) |

Where y is the observation, c is constant (intercepts), $\phi$ is coefficient, and $\epsilon$ is error item.

### B. Train and Test Data Split

In ML, the dataset is commonly split into two sets: training and testing. For our model training, we used the year of 2019 [1/1/2019, 12/31/2019] as our training dataset, and one month of 01/2020 [1/1/2020, 1/31/2020] as our testing dataset.

### C. LAS Departure Demand Forecasting Results

Table 4 compares five models' performances with different combinations of data sources and modeling methods. The best evaluation metrics (mse, mae, and explained variance) are highlighted in yellow. For mse and mae values, a lower value is better and for the explained variance score, a higher value is better. It is obvious that seq2seq with attention, combined with input data sources of ASPM and SWIM, achieves the best performance. Its mse is 62% better than the AR model, and 39% better than the LR model. Since SWIM data are updated every minute, we can shorten the look ahead times from 31 hours to 4 hours, and at the same time update our predictions every 15 minutes. Even when a single data source (ASPM) is used, seq2seq with attention still produces the predictions with the lowest error.

**Table 4 Model performances comparison**

| data | model | mse | mae | explained_variance | n_lag | n_look_ahead | mse_comparison |
|---|---|---|---|---|---|---|---|
| ASPM | Linear_Regression | 7.63 | 2.1 | 0.65 | 10 | 124 | -39% |
| ASPM | Autoregressive | 8.91 | 2.3 | 0.59 | 96 | 124 | -62% |
| ASPM | Seq2Seq | 6.53 | 1.8 | 0.72 | 10 | 124 | -19% |
| ASPM | Seq2Seq_Attention | 6.27 | 1.8 | 0.72 | 10 | 124 | -14% |
| ASPM+SWIM | Seq2Seq_Attention | 5.49 | 1.7 | 0.75 | 10 | 8 | 0% |

Fig. 8 shows an example of LAS quarter-hour departure demand forecasting results by seq2seq with attention model for several testing days. The blue line represents the true departure demand, and the red line is the forecasted demand by seq2seq with attention model. From the graph, the seq2seq with attention model does a pretty good job predicting the actual departure demand, including capturing demand surges at some periods.

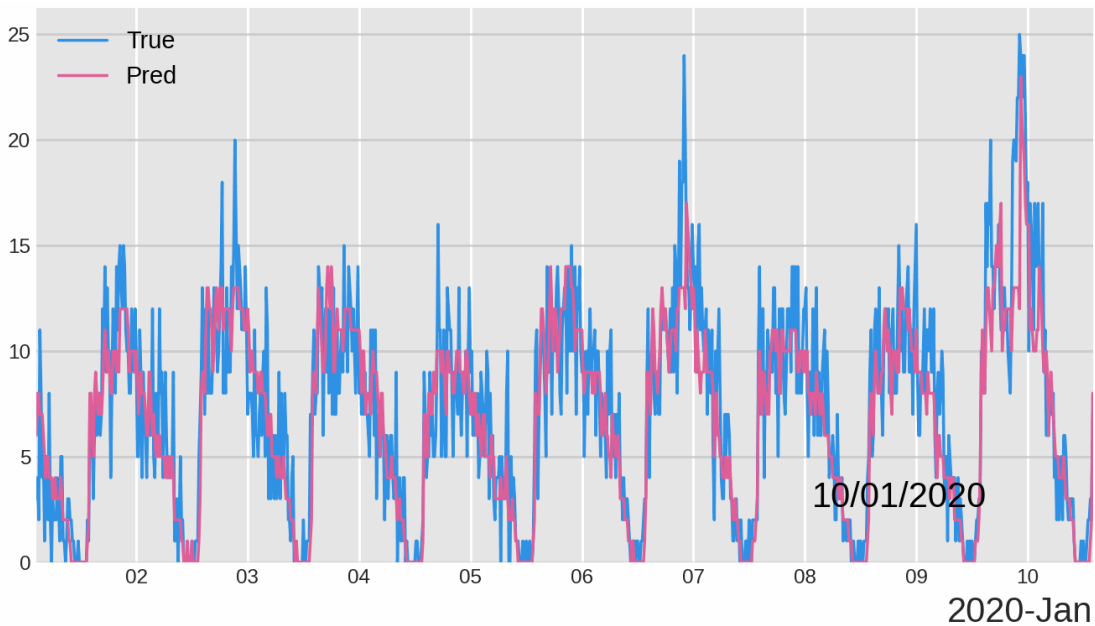

**Fig. 8 LAS quarter-hour departure demand forecasting by seq2seq with attention model**

Fig. 9 compares the five models' forecasted quarter-hour departure demand at LAS airport with true demand (red). According to Fig. 9, only seq2seq with attention model combining SWIM near live data can capture the sudden demand surge on 1/10/2020.

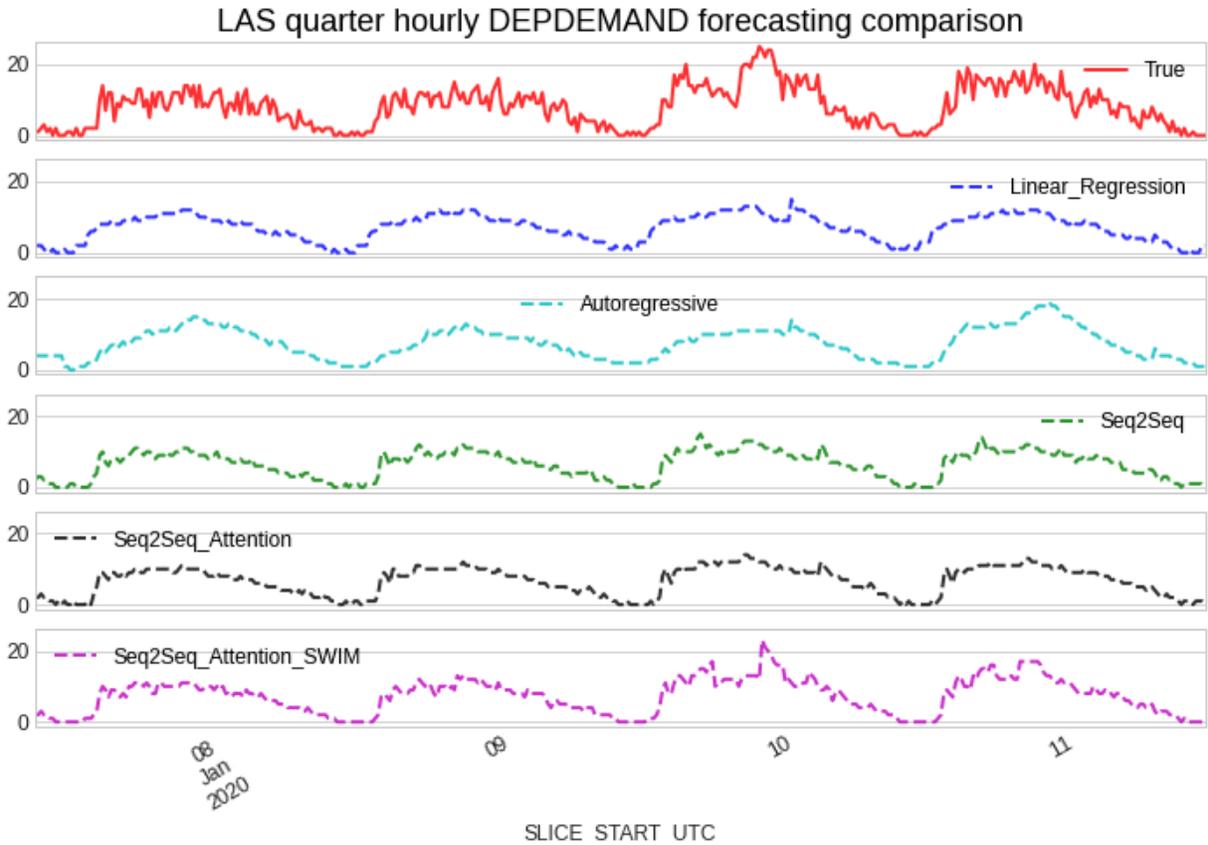

**Fig. 9 LAS models quarter-hour departure demand forecasting results comparison**

Additionally, we aggregated the quarter-hour forecasted demand to hourly and daily forecasts. Fig. 10 and Fig. 11 show the comparison of the hourly and daily forecasting results. According to the evaluation metrics, seq2seq with attention combining SWIM data still performs the best in those two cases. In comparison, AR still performs the worst over all three-evaluation metrics.

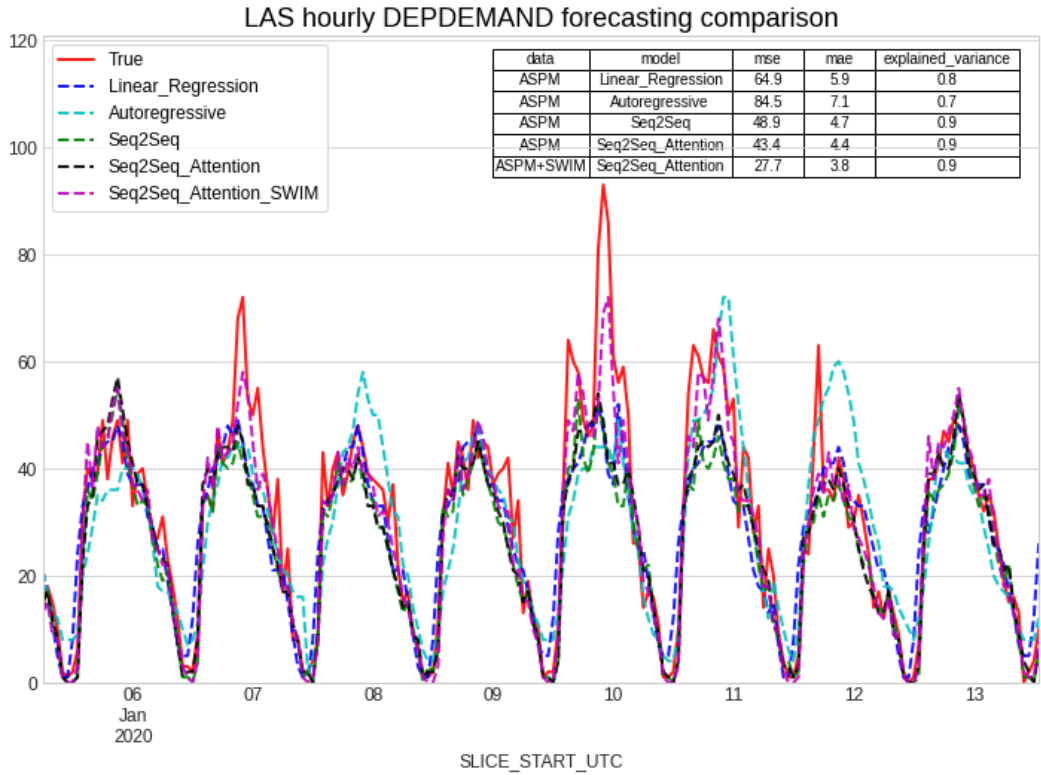

**Fig. 10 LAS hourly departure demand forecasting results comparison**

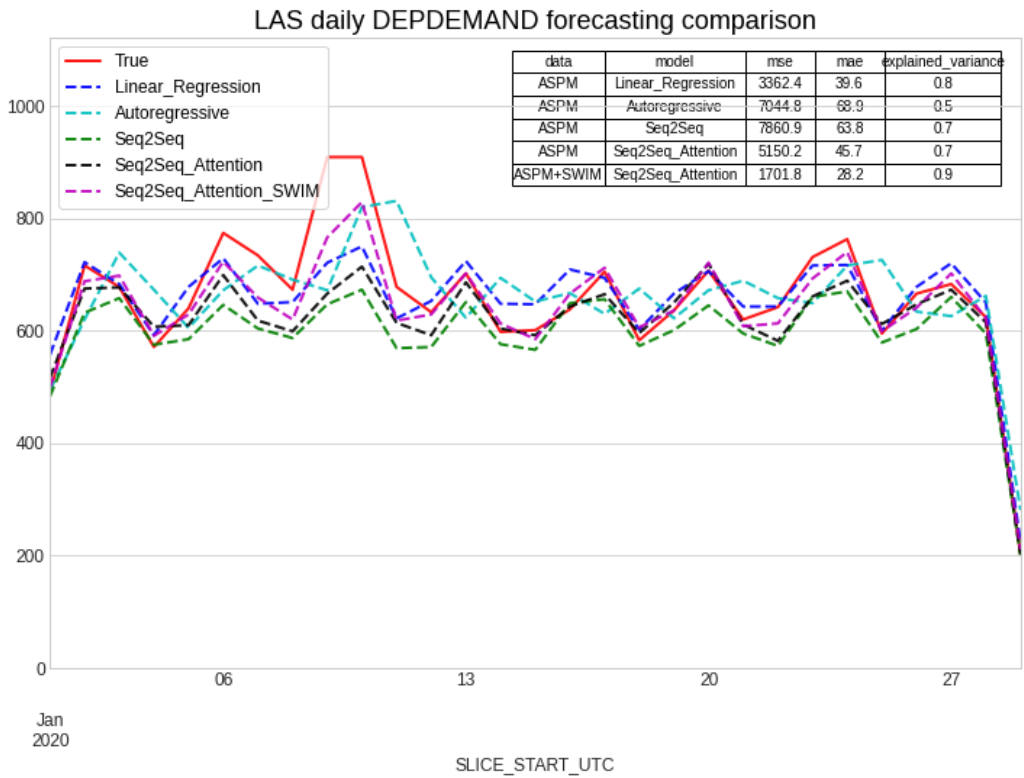

**Fig. 11 LAS daily departure demand forecasting results comparison**

In short, the LAS case study demonstrates superiority of the DL method, seq2seq with attention combined with SWIM data over the other examined methods.

## VII. Conclusions and Future Work

This paper applied DL models, seq2seq and seq2seq with attention, to improve flight departure demand forecast accuracy for Pacer, which aims to improve surface situation awareness and operational predictability at airports with a sizeable number of GA flight operations that experience departure delays and surface congestion. The research has demonstrated that combining good input data and robust forecasting algorithms is a key to obtaining the best prediction performance. With both ASPM and SWIM as input data sources, the model can reduce mse by 62% over the baseline algorithm of AR. Currently, we are also exploring the use of a transformer model for departure demand forecasting to seek even better performance.

**Acknowledgments**

We thank the following MITRE colleagues: Paul Diffenderfer, Kevin Long, Joey Menzenski, Dr. Ronald Chong, Dr. Travis Gaydos, Diane Baumgartner, Caroline Abramson, Dr. Emily Stelzer, Dr. Erik Vargo, Dr. Alex Tien, Brennan Haltli, Dr. Panta Lucic, and Suzanne Porter for their valuable discussions and insights.

**References**

[1] MITRE, "Pacer – Departure Readiness," [Online]. Available: https://sites.mitre.org/mobileaviationresearch/pacer-original/. [Accessed 22 Oct 2020].

[2] I. Sutskever, O. Vinyals and Q. V. Le, "Sequence to Sequence Learning with Neural Networks," in *In Advances in Neural Information Processing Systems*, Montréal, Canada, 2014.

[3] M.-T. Luong, H. Pham and C. D. Manning, "Effective Approaches to Attention-based Neural Machine Translation," 20 September 2015. [Online]. Available: https://arxiv.org/pdf/1508.04025.pdf. [Accessed 20 October 2020].

[4] K. C. Y. B. Dzmitry Bahdanau, "Neural Machine Translation by Jointly Learning to Align and Translate," in *3rd International Conference on Learning Representations (ICLR)*, San Diego, CA, 2015.

[5] Z. Wang, S. Lance and J. Shortle, "Improving the Nowcast of Unstable Approaches," in *Integrated Communications, Navigation, Surveillance (ICNS)*, Dullas,VA, 2016.

[6] A. Maheshwari, N. Davendralingam and D. DeLaurentis, "A Comparative Study of Machine Learning Techniques forAviation Applications," in *Aviation Technology, Integration, and Operations Conference*, Altalanta, GA, 2018.

[7] Y. J. Kim, S. Choi, S. Briceno and D. Mavris, "A deep learning approach to flight delay prediction," in *IEEE/AIAA 35th Digital Avionics Systems Conference (DASC)*, Sacramento, CA, 2016.

[8] H. Lee and W. Malik, "Taxi-Out Time Prediction for Departures at Charlotte Airport Using Machine Learning Techniques," in *16th AIAA Aviation Technology, Integration, and Operations Conference*, Washington, D.C., 2016.

[9] FAA, "ASPM," [Online]. Available: https://aspmhelp.faa.gov/index.php/Aviation_Performance_Metrics_%28APM%29. [Accessed 26 May 2020].

[10] FAA, "ASPM Airport Quarter Hour Data Dictionary," FAA, [Online]. Available: https://aspm.faa.gov/aspm/Dict_AirportQtr.pdf. [Accessed 28 May 2020].

[11] FAA, "SWIM," FAA, [Online]. Available: https://www.faa.gov/air_traffic/technology/swim/overview/. [Accessed 10 June 2021].

[12] FAA, "Core 30 Airports," [Online]. Available: https://aspm.faa.gov/aspmhelp/index/Core_30.html. [Accessed 10 June 2021].

[13] B. Lim, S. O. Arik, N. Loeff and T. Pfister, "Temporal Fusion Transformers for Interpretable Multi-horizon Time Series Forecasting," ArXiv, 27 September 2020. [Online]. Available: https://arxiv.org/abs/1912.09363. [Accessed 12 January 2021].

[14] J. Brownlee, "How to Develop Multivariate Multi-Step Time Series Forecasting Models for Air Pollution," [Online]. Available: https://machinelearningmastery.com/how-to-develop-machine-learning-models-for-multivariate-multi-step-air-pollution-time-series-forecasting/. [Accessed 1 May 2020].

[15] K. Cho, B. V. Merrienboer, C. Gulcehre, D. Bahdanau, F. Bougares, H. Schwenk and Y. Bengio, "Learning Phrase Representations using RNN Encoder–Decoder for Statistical Machine Translation," in *Empirical Methods in Natural Language Processing (EMNLP)*, Doha, Qatar, 2014.

[16] W. Wang, "The Amazing Effectiveness of Sequence to Sequence Model for Time Series," [Online]. Available: https://weiminwang.blog/2017/09/29/multivariate-time-series-forecast-using-seq2seq-in-tensorflow/. [Accessed 1 May 2020].

[17] S. Madhu, "Chapter 10: DeepNLP - Recurrent Neural Networks with Math," 10 Jan 2018. [Online]. Available: https://medium.com/deep-math-machine-learning-ai/chapter-10-deepnlp-recurrent-neural-networks-with-math-c4a6846a50a2. [Accessed 8 Aug 2020].

[18] Wiki, "Linear Regression," [Online]. Available: https://en.wikipedia.org/wiki/Linear_regression#cite_note-Freedman09-1. [Accessed 15 Aug 2020].

[19] Wiki, "Vector autoregression," [Online]. Available: https://en.wikipedia.org/wiki/Autoregressive_model. [Accessed 1 June 2020].